\documentclass[10pt,twocolumn,letterpaper]{article}
\usepackage{cvpr}
\usepackage{times}
\usepackage{epsfig}
\usepackage{graphicx}
\usepackage{amsmath}
\usepackage{amssymb}

\usepackage{multirow}
\usepackage{booktabs} 

\usepackage{authblk}
\usepackage[misc]{ifsym}

\usepackage{enumitem}
\usepackage{multirow}
\usepackage{diagbox} 
\newcommand{\mypar}[1]{{\bf #1.}}

\newcommand{\R}{\ensuremath{\mathbb{R}}}

\def\h{\mathbf{h}}

\def\dd{\mathbf{d}}
\def\e{\mathbf{e}}
\def\p{\mathbf{p}}
\def\q{\mathbf{q}}
\def\vv{\mathbf{v}}

\def\z{\mathbf{z}}
\DeclareMathOperator{\Ss}{S}
\DeclareMathOperator{\Pj}{P}

\DeclareMathOperator{\Z}{Z}

\usepackage[pagebackref=true,breaklinks=true,letterpaper=true,colorlinks,bookmarks=false]{hyperref}

\cvprfinalcopy 

\def\cvprPaperID{3100} 

\ifcvprfinal\fi
\begin{document}

\title{Collaborative Motion Prediction via Neural Motion Message Passing}

\author[1]{Yue Hu}
\author[2 \Letter]{Siheng Chen}
\author[1 \Letter]{Ya Zhang}
\author[1]{Xiao Gu}

\affil[1]{{ Cooperative Medianet Innovation Center, Shanghai Jiao Tong University}}
\affil[2]{{ Mitsubishi Electric Research Laboratories}
\authorcr {\tt\small \{18671129361, ya\_zhang, gugu97\} @sjtu.edu.cn, schen@merl.com}}


\maketitle
\thispagestyle{empty}

\begin{abstract}
Motion prediction is essential and challenging for autonomous vehicles and social robots. One challenge of motion prediction is to model the interaction among traffic actors, which could cooperate with each other to avoid collisions or form groups. To address this challenge, we propose neural motion message passing (NMMP) to explicitly model the interaction and learn representations for directed interactions between actors. Based on the proposed NMMP, we design the motion prediction systems for two settings: the pedestrian setting and the joint pedestrian and vehicle setting. Both systems share a common pattern: we use an individual branch to model the behavior of a single actor and an interactive branch to model the interaction between actors, while with different wrappers to handle the varied input formats and characteristics. The experimental results show that both systems outperform the previous state-of-the-art methods on several existing benchmarks. Besides, we provide interpretability for interaction learning. Code is available at~\url{https://github.com/PhyllisH/NMMP}.

\end{abstract}

\vspace{-0.1in}
\section{Introduction}
Forecasting the future motion for the interacting actors in the scene has been a crucial problem in many real-world scenarios. For example, self-driving vehicles and interactive robotics need to understand human and other traffic actors' future behaviors to avoid collisions and for better planning~\cite{casas2018intentnet,chou2019predicting,djuric2018short,forestier2016autonomous}. The intelligent tracking modules in surveillance systems also need to understand the pedestrians' motion to optimize resource allocation~\cite{sultani2018real}.  Scientifically, motion prediction is also useful for understanding human behaviors~\cite{li2019actional} and motion dynamics~\cite{qi2018learning}.

\begin{figure}[t]
\begin{center}
\centerline{\includegraphics[width=0.5\linewidth]{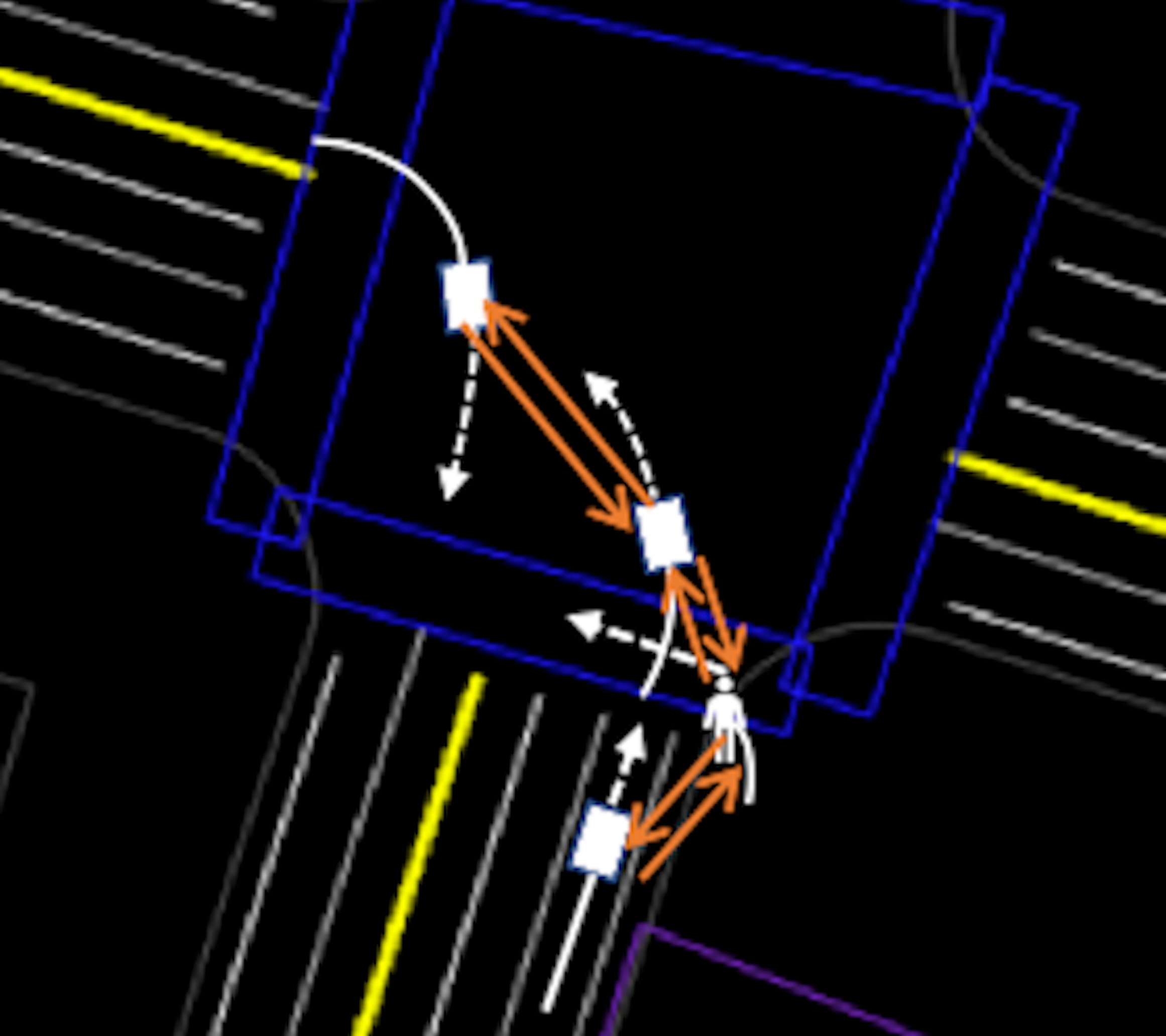}}
\caption{Neural motion massage passing is proposed to capture directed interactions among traffic actors, which can serve as a plugin module to guide motion prediction.}
\label{fig:crown_jewl}
\end{center}
\vspace{-0.4in}
\end{figure}

One of the fundamental challenges of this task is to model the scene constraints, especially the hidden interactions between actors. For example, in the driving scenarios, traffic actors, such as vehicles and pedestrians, are influenced by each other, as well as the traffic conditions and rules; see Figure~\ref{fig:crown_jewl}. Previous works have tried to model such interactions through three mechanisms: spatial-centric mechanism, social mechanism, and graph-based mechanism. The spatial-centric mechanism uses the spatial relationship to implicitly model the interaction between actors~\cite{deo2018convolutional,zhao2019multi,bansal2018chauffeurnet,djuric2018short,chou2019predicting,casas2018intentnet}. The social mechanism explicitly aggregates the neighboring actors' information involved in the scene through social pooling~\cite{alahi2016social,gupta2018social}, or attention operation~\cite{vemula2018social,sadeghian2019sophie}; however, both mechanisms model the interactions according to the static spatial locations of the actors and ignore the temporal horizon. The graph-based mechanism constructs a graph to explicitly model the pairwise interactions between actors according to their observed trajectories~\cite{kosaraju2019social, vemula2018social, huangstgat}; however, the previous graph-based methods only consider features for actors, but do not explicitly learn features for in-between interactions.

To comprehensively represent the interactions between actors, we extend the current graph-based mechanism and propose neural motion message passing (NMMP). The NMMP infers an interaction graph from actors' trajectories, whose nodes are the actors and edges are the interactions. This module updates the neural message passing framework~\cite{kipf2018neural} to the context of motion prediction. It takes the observed trajectories of traffic actors as input, and produces the feature representations for both actors and their corresponding interactions. A main advantage of the proposed NMMP over the current graph-based methods is that NMMP not only presents informative feature representations for both actors and interactions, but also provides interpretability for the motion prediction task. Interaction representations indicating similar interaction patterns have similar impacts on the acting actors explicitly. Besides, we consider the direction of interaction.

We design the motion-prediction systems for two specific settings: the pedestrian setting, where we consider an open area and the actors are flexible pedestrian; and the joint pedestrian and vehicle setting, where we consider an urban driving scenario and an actor could be either pedestrian or vehicle. Most previous literature focus on only one of these two settings. Here we handle both settings based on the same NMMP module to model interactions, and an individual branch to model the behavior of a single actor and an interactive branch to model the interaction between actors. We handle the different input formats and characteristics in these two settings with distinct wrappers.
The pedestrian motion prediction system includes additional GAN to deal with the uncertain human behaviours in open-area scenes. The joint motion prediction system includes additional CNN to process map information required in urban-driving scenes, and coordinate transformation to handle vehicle headings. Overall, those differences between two systems are small variations. The experimental results show that the proposed systems outperform the state-of-the-art methods in both settings, which not only show the superiority of the proposed systems, but also validate the generalization and interpretability of the proposed NMMP. The main contributions of this paper are:
\vspace{-2mm}
\begin{itemize}
    \item We propose neural motion message passing (NMMP) to learn feature representations for both actors and their interactions from actors' trajectory;
\vspace{-2mm}
    \item We design systems for pedestrian motion prediction,  and joint pedestrian and vehicle motion prediction based on the NMMP module; both outperform the previous state-of-the-art methods;
\vspace{-2mm}
    \item We provide interpretability for the proposed motion prediction systems, including both quantitative benchmarks and visualization analysis.
\end{itemize}

\vspace{-2mm}
\section{Related Works}
\vspace{-2mm}
\textbf{Motion prediction.} Traditional approaches for motion prediction are based on hand-crafted rules and energy potentials~\cite{antonini2006discrete,lee2007trajectory,mehran2009abnormal,morris2009learning,wang2011trajectory,wang2008unsupervised}. For example, Social Force~\cite{helbing1995social} models pedestrian behavior with attractive and repulsive forces; however, those hand-crafted features fail to generalize to complex scenes. To tackle this problem, researchers tend towards data-driven tools. For example, the sequence-to-sequence models, such as recurrent neural networks~\cite{sutskever2014sequence}, are leveraged to encode prior trajectory sequences~\cite{alahi2016social, lee2017desire}; however, those models consider the behavior of each individual actor and ignore the rich interactions among multiple actors in a scene. Recently, three mechanisms are developed to model the hidden interactions.

The first one is the~\emph{spatial-centric mechanism}; it represents the actors' trajectories in a unifying spatial domain and uses the spatial relationship to implicitly model the interaction between actors. For example, Social Conv~\cite{deo2018convolutional} and MATF~\cite{zhao2019multi} leverage the spatial structure of the actors to learn the interactions; ChauffeurNet~\cite{bansal2018chauffeurnet} and Motion Prediction~\cite{djuric2018short} encode the trajectories of traffic actors and the scene context into bird's eye view images;  FMNet~\cite{chou2019predicting} uses lightweight CNNs to achieve the real-time inference; and IntentNet~\cite{casas2018intentnet} combines LiDAR data with images.

The second one is the~\emph{social mechanism}; it aggregates the neighboring actors' information to a social representation and broadcasts it to each actor. In this way, each actor is aware of the neighboring information.  For example, Social LSTM~\cite{alahi2016social} utilizes max pooling over neighboring actors. To consider long-range interactions, Social GAN~\cite{gupta2018social} applies max pooling to all the actors. CIDNN~\cite{xu2018encoding} uses inner product between the prior location embeddings of actors. However, the max-pooling operation ignores the uniqueness of each actor and the inner-product operation considers all the other actors equally. The attention operation is then introduced~\cite{vemula2018social,sadeghian2019sophie} so that the actor could focus on crucial impacts. Inevitably, increasing computational complexity comes along with the attention operation. 

The third one is the~\emph{graph-based mechanism}; it constructs a graph to explicitly model the pairwise interactions between actors. For example, Social-BiGAT\cite{kosaraju2019social} learns a global embedding to represent the interactions in the scene based on a graph attention network(GAT); Social Attention~\cite{vemula2018social} and STGAT~\cite{huangstgat} capture the dynamic interaction changes over time by using  
spatial-temporal graphs and LSTM, respectively. In this work, we extend the graph-based mechanism from two aspect: (i) capture directed interactions; and (ii) provide interpretability for interactions. 

\textbf{Graph neural network.} Graph neural networks recently have got a lot of attention and achieved significant success in various fields~\cite{gilmer2017neural, niu2018generalized, kipf2018neural}, especially in social network analysis~\cite{hamilton2017inductive}, scene understanding~\cite{woo2018linknet,yang2018graph}, 3D point cloud processing~\cite{li2019deepgcns} and human action understanding~\cite{qi2018learning, li2019actional}. Two mainstream architectures include graph convolutional networks~\cite{scarselli2008graph, li2015gated, velivckovic2017graph} and neural-message-passing-based networks~\cite{gilmer2017neural, kipf2018neural}. While graph convolutional networks consider the edges as a transient part, neural-message-passing-based networks treat them as a integral part of the model. In this work, we use the neural-message-passing-based networks to learn complicated interactions between traffic actors, where actors are considered as nodes and interactions are considered as edges.

\vspace{-2mm}
\section{Neural Motion Message Passing}
\vspace{-1mm}
In this section, we present NMMP, whose functionality is to provide feature representations for both actors and their corresponding interactions from the input trajectories. This NMMP serves as a core module in the proposed systems as it enables the traffic actors to share their history information and collaboratively predict their future trajectories.

Considering $N$ visible traffic actors within the scene. With $\p_i^{(t)} = (x_i^{(t)},y_i^{(t)})\in \R^2$ the spatial coordinate of the $i$th actor at timestamp $t$, let the observed trajectory $\Pj^-_i$ and the ground-truth future trajectory $\Pj^+_i$ of the $i$th actors be 
\begin{align*}
\vspace{-2.5mm}
  \Pj^-_i \ &= \   
  \begin{bmatrix}
    \p^{(-T_{\rm obs})}_{i} & \p^{(-T_{\rm obs}+1)}_{i} & \ldots  &  \p^{(0)}_{i}
  \end{bmatrix} \ \in \ \R^{2 \times (T_{\rm obs+1})}, \\
\vspace{-1mm}
\vspace{-1mm}
  \Pj^+_i \ &= \   
  \begin{bmatrix}
    \p^{(1)}_{i} & \p^{(2)}_{i} & \ldots  &  \p^{(T_{\rm pred})}_{i}
  \end{bmatrix} \ \in \ \R^{2 \times T_{\rm pred}}.
  \vspace{-0.5mm}
\end{align*}
The overall goal of motion prediction is to propose a prediction model $g(\cdot)$, so that the predicted future trajectories
$
 \{ \widehat{\Pj}^+_i \}_{i=1}^N  \ = g\bigg( \{ \Pj^-_i \}_{i=1}^N \bigg)
$
are close to the ground-truth future trajectories
$\{ {\Pj}^+_i \}_{i=1}^N$. 

\begin{figure}[tp]
\begin{center}
\centerline{\includegraphics[width=0.95\linewidth]{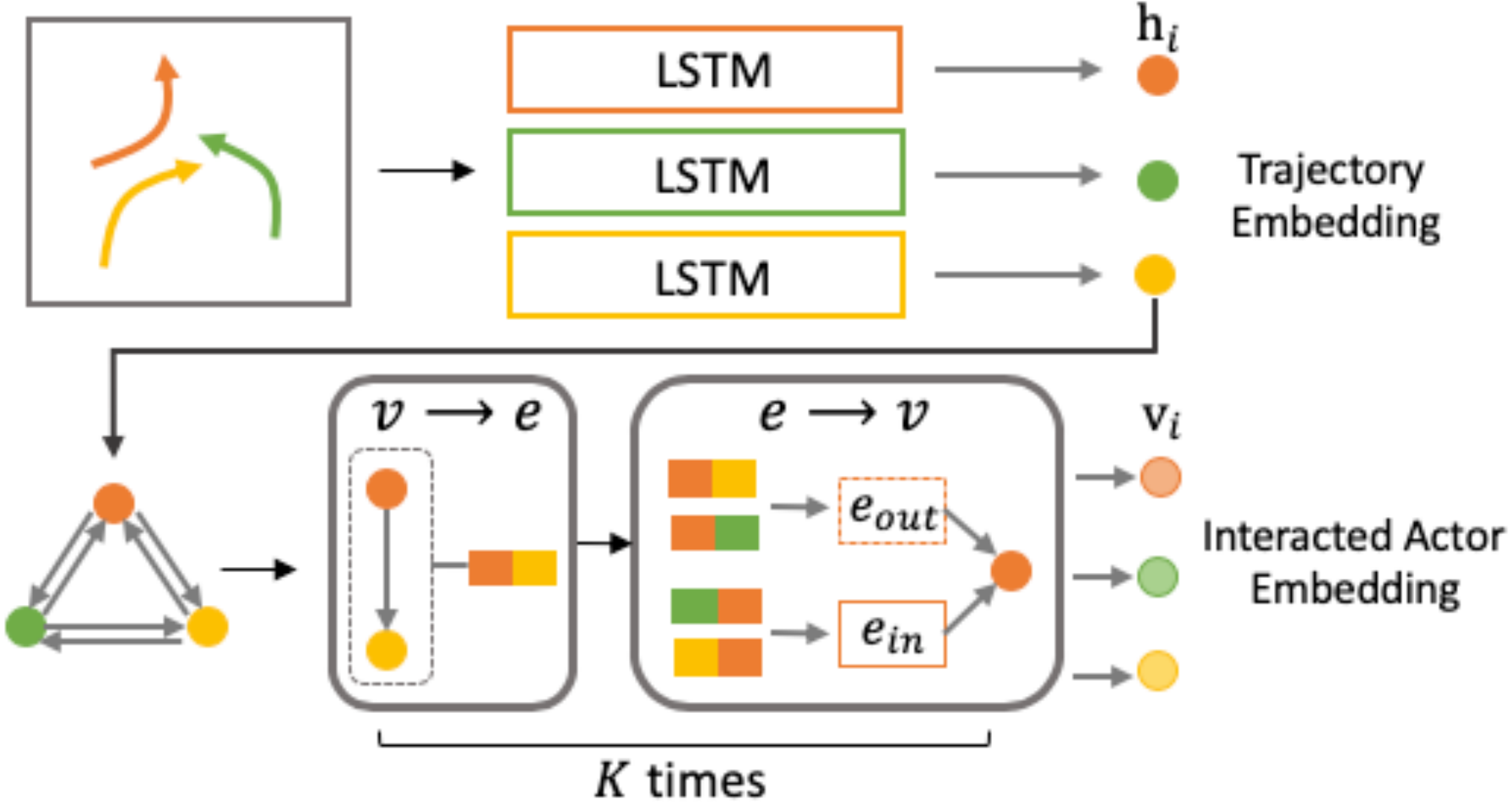}}
\caption{The NMMP module explicitly learns trajectory embedding for each actor through LSTM, and interacted actor embedding and in-between interaction embedding through message passing.}
\label{fig: NMMP}
\end{center}
\vspace{-0.4in}
\end{figure}

See Figure~\ref{fig: NMMP}, intuitively, the actors are interactively influenced by each other in real-time, leading to correlated trajectories. The NMMP is proposed to leverage such a correlation to provide informative feature representations for the actors and their corresponding interactions. The core of NMMP is an interaction graph, $G(\mathcal{V}, \mathcal{E})$, where the node $v_i \in \mathcal{V}$ represents the $i$th traffic actor, and the edge $e_{ij} \in \mathcal{E}$ denotes the interaction between two actors, $v_i$ and $v_j$. We initialize it as a fully-connected graph and then initialize the actor and interaction embedding with trajectory embedding which is obtained though
\begin{subequations}
\label{eq:traj_emb}
\begin{eqnarray}
\vspace{-3mm}
\label{eq:map_displacement}
\h_i^{(t)} & = & f_{\rm temp}(\p_i^{(t)}-\p_i^{(t-1)}),
\\
\label{eq:nmp_lstm}
\h_i & = & f_{\rm LSTM}( \{ \h_i^{(t)} \}_{t = -T_{\rm obs}+1}^{0} ) \in \R^D, 
\\
\label{eq:nmp_init_spatial}
\dd_{ij} & = & f_{\rm spatial} \left( \p_i^{(0)}-\p_j^{(0)} \right).
\vspace{-3mm}
\end{eqnarray}
\end{subequations}
we encode the displacement between the coordinates at two consecutive timestamps in~\eqref{eq:map_displacement}. The LSTM is utilized to integrate the temporal information to obtain the trajectory embedding $\h_i$ of the $i$th actor in~\eqref{eq:nmp_lstm}. And~\eqref{eq:nmp_init_spatial} encodes the difference between the actors at the current time, providing the relative spatial information.
Then, map the trajectory embedding to the actor space, we get the initial actor embedding of the $i$th actor, $\vv_i^{0}=f_v^0  \left(\h_i\right)$, and concatenate both actor embeddings and the relative spatial embedding $\dd_{ij}$, we get the initial interaction embedding between the $i$th and the $j$th actors, $\e_{ij}^{0}=f_e^0 \left([\vv_i^{0}; \vv_j^{0}; \dd_{ij}]\right)$, which includes both temporal and spatial information. 
$f_{\rm temp}(\cdot)$, $f_{\rm LSTM}(\cdot)$, $f_{\rm spatial}(\cdot)$, $f_v^0(\cdot)$, and $f_e^0(\cdot)$ are MLPs.


\begin{figure*}[!htp]
\vspace{-2mm}
\begin{center}
\centerline{\includegraphics[width=0.9\textwidth]{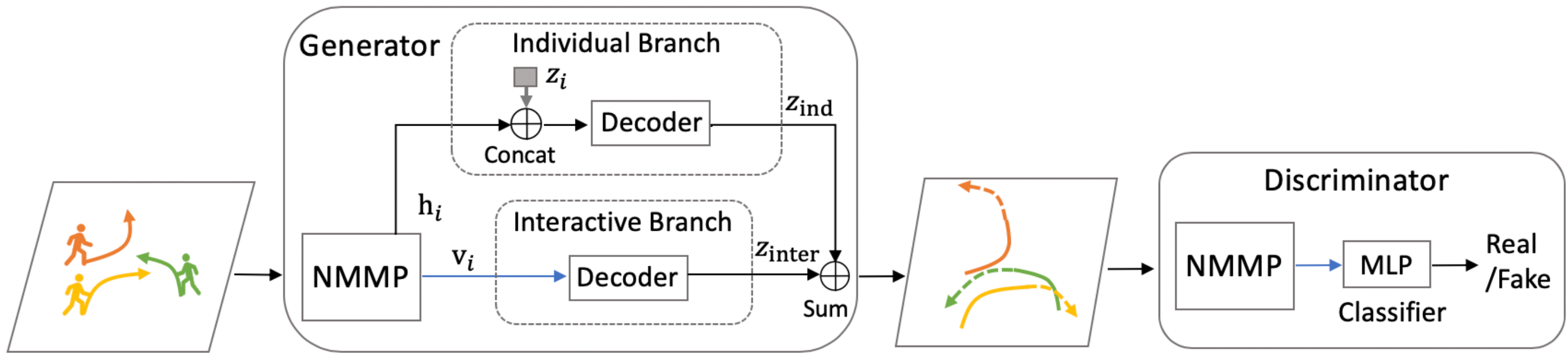}}
\vspace{-1mm}
\caption{The pedestrian motion prediction system based on NMMP (PMP-NMMP) consists of a generator and a discriminator. The generator predicts future trajectories from two branches: the individual branch considers the individual trajectory embedding and the interactive branch considers the interacted actor embedding. The discriminator classifies the complete trajectories to be real or fake.}
\label{fig: framework}
\end{center}
\vspace{-10mm}
\end{figure*}

Following the spirit of the node-to-edge and edge-to-node message passing mechanism~\cite{gilmer2017neural, kipf2018neural},
we update the interaction graphs with the associated actor and interaction embeddings; see Figure~\ref{fig: NMMP}. In the node-to-edge phase, each interaction receives motion messages from the corresponding pair of actors. In the edge-to-node phase, each actor embedding is updated according to all the corresponding interaction embeddings. The $k$th ($k \geq 0$) iteration works as
\begin{subequations}
\begin{eqnarray}
\label{eq:edge2node} 
\vv_i^{k+1} & = & f_v^{k+1}([\frac{1}{d_i^{\rm in}}\sum\limits_{e_{ji}\in \mathcal{E}}\e_{ji}^{k} ; \frac{1}{d_i^{\rm out}}\sum\limits_{e_{ij}\in \mathcal{E}}\e_{ij}^{k}]),
\\
\label{eq:node2edge} 
\e_{ij}^{k+1} & = & f_e^{k+1}\left([\vv_i^{k+1}; \vv_j^{k+1}]\right), 
\vspace{-3mm}
\end{eqnarray}
\end{subequations}
where ${f_e^{k}}(\cdot)$, ${f_v^{k}}(\cdot)$ are MLPs. $\vv_i^{k}$ and $\e_{ij}^{k}$ are the $i$th actor embedding and the interaction embedding between the $i$th and $j$th actors at the $k$th iteration, respectively.

Here we consider a directed graph to model the relative interactions between traffic actors. For example, when the front actor slows down, the back actor would be forced to slow down or turn to avoid collision, while inversely, the back actor slowing down might not equally affect the front actor's behavior. 
To explicitly reflect this directed information, in the edge-to-node phase~\eqref{eq:edge2node}, we use the concatenation rather than the sum or mean to distinguish the edge direction. $d_i^{\rm in}$ is the amount of edges pointing to $v_i$, while $d_i^{\rm out}$ is the amount of edges $v_i$ pointing out; both of which are used to normalize the interaction embeddings. In the node-to-edge phase~\eqref{eq:node2edge}, we concatenate two actor embeddings to update the interaction embedding. Repeat the node-to-edge and edge-to-node phases $K$ times, we can get the final interacted actor embeddings ($\vv_i = \vv_i^{K}$) and the final interaction embeddings ($\e_{ij} = \e_{ij}^{K}$). The interaction density is positively related to the number of iterations $K$.

\vspace{-1mm}
\section{Motion Prediction Systems}
\vspace{-1mm}
\label{section: method}
We propose the motion-prediction systems based on NMMP for two settings: pedestrian motion prediction, and joint pedestrian and vehicle motion prediction.
\vspace{-1mm}
\subsection{Pedestrian motion prediction}
\vspace{-1mm}
The pedestrian motion prediction considers open areas, such as campus squares, where pedestrians walk flexibly. The trait of setting is the trajectory is highly nonsmooth. This setting fits to the scenarios of surveillance systems.

\vspace{-3mm}
\subsubsection{System architecture}
\vspace{-1mm}
This system consists of a generator and a discriminator; see Figure~\ref{fig: framework}. 
The generator predicts future trajectories of the actors and the discriminator classifies the complete trajectories to be real or fake. The model is trained adversarially to encourage realistic trajectories. 

\mypar{Generator}
We predict the future trajectory based on two branches: the individual branch, which provides a rough prediction based on each individual actor, and the interactive branch, which refines the rough prediction based on interaction information. The predicted spatial coordinate of the $i$th actor at timestamp $t$, $\widehat{\p}_i^{(t)}$, is obtained through
\begin{subequations}
\begin{eqnarray}
\vspace{-2mm}
\label{eq:pmp_local}
\z^{(t)}_{\rm ind} & = & g_{\rm ind}\left(g_{\rm LSTM}\left(\q_i^{(t)}, \widehat{\p}_i^{(t-1)} \right)\right) \in \R^2,
\\ 
\label{eq:pmp_context}
\z^{(t)}_{\rm inter} & = & g^{(t)}_{\rm inter} (\vv_i) \in \R^2,
\\
\label{eq:pmp_sum}
\widehat{\p}_i^{(t)} & = & \widehat{\p}_i^{(t-1)} + \z^{(t)}_{\rm ind} + \z^{(t)}_{\rm inter} \in \R^2,
\vspace{-2mm}
\end{eqnarray}
\end{subequations}
where $g_{\rm ind}(\cdot)$ and $g^{(t)}_{\rm inter}(\cdot)$ are MLPs. $\q_i^{(t)}$ is the hidden state of the $i$-th actor's LSTM at time $t$, which is intialized with $\q_i^{(0)} = [\h_i; \mathbf{z}_i]$, $\h_i$ from NMMP~\eqref{eq:nmp_lstm}, $\mathbf{z}_i$ is gaussian noise to encourage diversity. The LSTM $g_{\rm LSTM}(\cdot)$ predicts the future movements in time order.~\eqref{eq:pmp_local} predicts the future trajectories of each individual actor based on its observed trajectories;~\eqref{eq:pmp_context} predicts the interaction component with the interacted actor embedding $\vv_i$; and~\eqref{eq:pmp_sum} provides the final predicted coordinate, which is the sum of the predicted coordinate at the previous time stamp, and the predicted individual and interaction components. Note that instead of using the absolute location, we predict the displacement to previous moment, $\widehat{\p}_i^{(t)} - \widehat{\p}_i^{(t-1)}$, which generalizes better.

\mypar{Discriminator}
The discriminator classifies a complete trajectory to be real or fake. It uses an individual NMMP module followed by a classifier. For the ground-truth samples, the complete trajectory is $[\Pj^-_i;\Pj^+_i]$, which should be classified as real; for the generated samples, the complete trajectory is $[\Pj^-_i;\widehat{\Pj}^+_i]$, which should be classified as fake. The probability to be real is obtained as
\vspace{-1mm}
$$
p_i=d_{\rm cls}(d_{\rm NMMP}(d_{\rm LSTM}(d_{\rm MLP}(  [\Pj^-_i;\widehat{\Pj}^+_i]   )))),
$$
where $d_{\rm MLP}(\cdot)$ denotes the MLP, $d_{\rm LSTM}(\cdot)$ is the LSTM to aggregate the temporal information, $d_{\rm NMMP}(\cdot)$ is the NMMP module, $d_{\rm cls}(\cdot)$ represents the classifier. 

\vspace{-2mm}
\subsubsection{Loss function}
\vspace{-1mm}
To train the model, we consider two losses for a scene: the generator loss $L_{\rm G}$ and the discriminator loss $L_{\rm D}$, 
\begin{eqnarray*}
L_{\rm G} &=& \sum_{i\in{1,2,...,N}} ||\widehat{\Pj}^+_i - \Pj^+_i||_2^2,
\\
L_{\rm D} &=& \sum\limits_{i\in{1,2,...,N}} {\rm log}(D([\Pj^-_i; \Pj^+_i])) \nonumber \\
& + & {\rm log}(1-D([\Pj^-_i;\widehat{\Pj}^+_i])),
\end{eqnarray*}
where $D(\cdot)$ produces the real probability of the complete trajectory generated by the classifier in discriminator. The generator and the discriminator play a min-max game to get more stochastic and realistic predictions.

\vspace{-1mm}
\subsection{Joint pedestrian and vehicle motion prediction}
\vspace{-1mm}
The joint pedestrian and vehicle prediction considers urban driving scenes, where both vehicles and pedestrians are involved. The trait of this setting is that we need to consider distinct motion patterns for vehicle and pedestrian, as well as complicated environmental information, such as drivable areas for vehicles. This setting fits to autonomous driving.

\begin{figure}[t]
\begin{center}
\centerline{\includegraphics[width=0.95\linewidth]{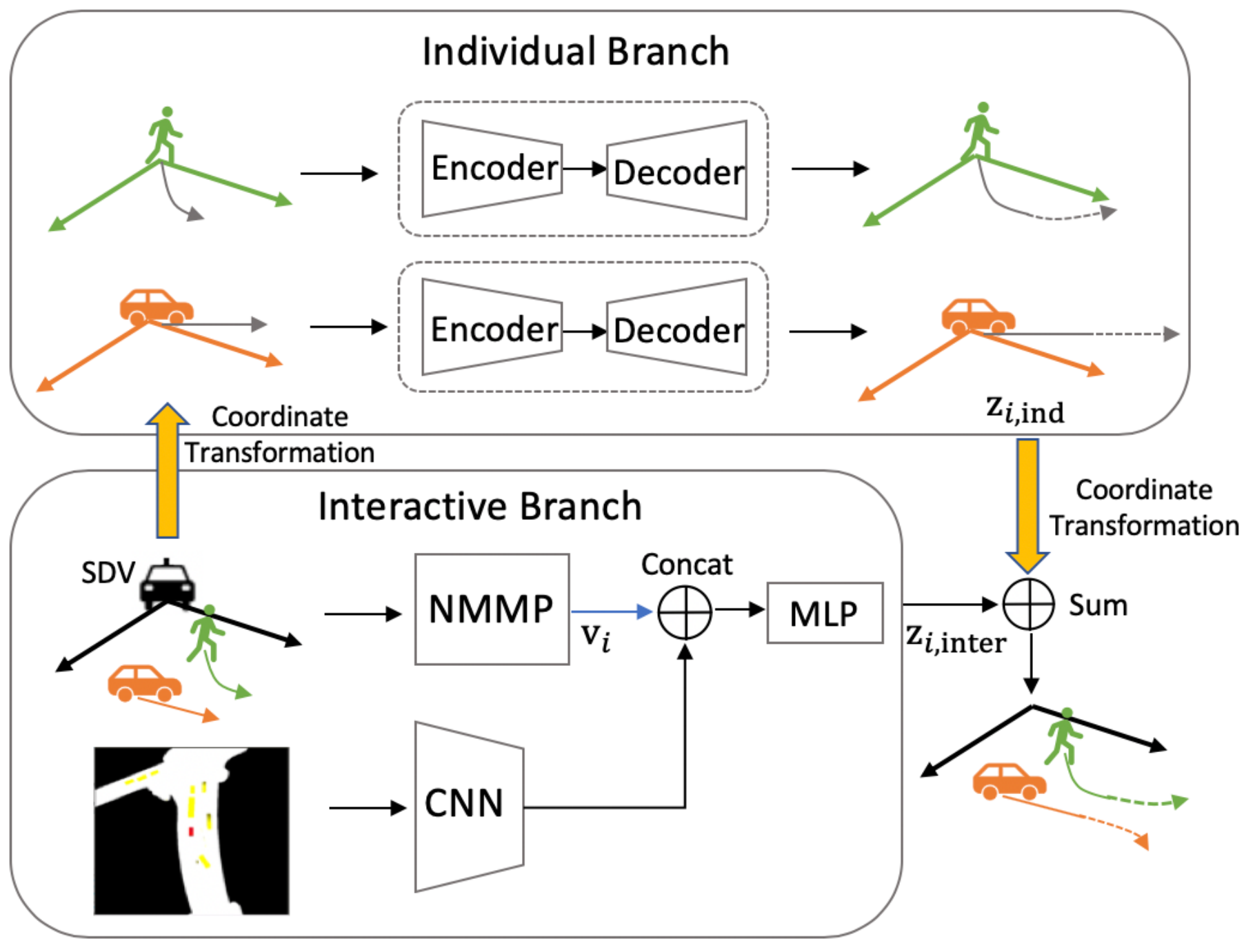}}
\caption{The joint pedestrian and vehicle motion prediction system based on NMMP (JMP-NMMP) predicts the future trajectories from two branches: the individual branch considers the individual behavior of each actor, and the interactive branch considers the interactions among actors.}
\label{fig: JointFramework}
\end{center}
\vspace{-0.4in}
\end{figure}

\vspace{-3mm}
\subsubsection{System architecture}
\vspace{-1mm}
This system includes an individual branches and an interactive branch; see Figure~\ref{fig: JointFramework}. Since vehicles and pedestrians have distinct motion behaviors, we use each individual branch for the same type of actors. The individual branch predicts the future trajectory for each actor without considering the interaction. The interactive branch predicts an additional compensation component by considering the interactions between actors. We then combine the outputs from both branches to obtain the final predicted trajectory.

\mypar{Individual branch} We predict the individual component of the future trajectory for each individual actor. We use ego-coordinate-system where the origin is the current position of each actor to reduce the influence of the starting position and heading and focus on the motion pattern. The individual branch follows encoder-decoder structure, where both encoder and decoder are implemented by MLPs. For the $i$th actor, the output of the individual branch is the individual component of the future trajectory, $\mathbf{Z}_{i, \rm ind} \in \R^{2 \times T_{\rm pred}}$, represented in its ego-coordinate-system.

\mypar{Interactive branch} We adopt the NMMP module to predict the interaction component of the future trajectory for each actor. Since we need to consider all the actors collaboratively, we use a global coordinate system. For example, we could consider the position of a self-driving vehicle (SDV) as the origin. We input all the actors' observed trajectories to the NMMP module, which considers the interactions and then outputs the interacted actor embedding $\vv_i$ for the $i$th actor. To further model the complicated environmental information, an additional 2D bird's-eye-view scene map is introduced. Such a rasterized scene image is important as it provides prior knowledge about the traffic condition and rules. Following the setting in~\cite{leal2014learning}, we rasterize the scene information to a RGB image; see the rasterization details in Appendix. We use MobileNet-V2~\cite{sandler2018inverted} to extract scene embedding, $\Ss$, from the RGB image. We then aggregate the interacted actor embeddding and the scene embedding through MLPs to obtain the output of the interactive branch. The interaction component of the future trajectory for the $i$th actor
$
  \Z_{i, {\rm inter}}  \ = \  {\rm MLP} \left( [\vv_i, \Ss] \right) \in \R^{2 \times T_{\rm pred}}.
$

Finally, we combine the outputs from both the individual branch and the interactive branch, and obtain the overall future trajectory.  The future trajectory for the $i$th actor is
\vspace{-0.05in}
\begin{equation*}
    \widehat{\Pj}^{+}_i \ = \ T_i\left(\Z_{i, \rm ind} \right) + \Z_{i, \rm inter} \in \R^{2 \times T_{\rm pred}},
\vspace{-0.05in}
\end{equation*}
where $T_i\left(\cdot\right)$ is the transform from the ego-coordinate-system of the $i$th actor to the global coordinate system.

\vspace{-2mm}
\subsubsection{Loss function}
\vspace{-1mm}
To train the system, we consider two losses: one for output from the individual branch and the other for the final combination; that is,
\begin{eqnarray*}
\vspace{-3mm}
L_{\rm ind} &=& \sum_{i\in{1,2,...,N}} ||T_i\left(\Z_{i, \rm ind}\right) - \Pj^{+}_i||_2^2,  \\
L_{\rm final} &=& \sum_{i\in{1,2,...,N}} ||\widehat{\Pj}^{+}_i - \Pj^{+}_i ||_2^2,
\vspace{-5mm}
\end{eqnarray*}
where $L_{\rm ind}$ is the $\ell_2$ loss between the individual component and the ground-truth future trajectory, and $L_{\rm final}$ is the $\ell_2$ loss between the final predictions and the ground-truth future trajectory. We obtain the final loss as,
$L =  \lambda L_{\rm ind} + (1 - \lambda) L_{\rm final}$. In the experiment, we set $\lambda = 0.5$.


\begin{table*}[ht]
	\caption{Quantitative results on~\emph{ETH-UCY} dataset. We present ADE/FDE in meters. Given previous 8 (3.2s), predicting future 12 (4.8s).}
	\label{tab:eth-ucy}
	\vskip 0.1in
	\begin{center}
	    \scriptsize
        \begin{footnotesize}
			\begin{sc}
				\begin{tabular}{c|c|c|c|c|c|c|c|c|c}
					\toprule 	
					 &\multicolumn{8}{c|}{Baselines} &\multicolumn{1}{c}{Ours} \\
					\midrule
					Dataset 
					&SLSTM &SAT &CIDNN &SGAN &Sophie &S-BiGAT &MATF GAN &STGAT &PMP-NMMP  \\
					\midrule
					ETH 
					&1.09/2.35 &1.39/2.39 &1.25/2.32 &0.87/1.62 &0.70/1.43 &0.69/1.29 &1.01/1.75  &0.65/1.12 &\textbf{0.61/1.08} \\
					HOTEL 
					&0.79/1.76 &2.51/2.91 &1.31/2.36 &0.67/1.37 & 0.76/1.67  &0.49/1.01 &0.43/0.80 &0.35/0.66 &\textbf{0.33/0.63}\\
					UNIV  
					&0.67/1.40 &1.25/2.54 &0.90/1.86 &0.76/1.52 & 0.54/1.24  &0.55/1.32 &\textbf{0.44/0.91} &0.52/1.10 &0.52/1.11\\
					ZARA1 
					&0.47/1.00 &1.01/2.17 & 0.50/1.04 &0.35/0.68 &0.30/0.63  &0.30/0.62 &\textbf{0.26/0.45} &0.34/0.69 &0.32/0.66\\
					ZARA2  
					&0.56/1.17 &0.88/1.75  & 0.51/1.07 &0.42/0.84 &0.38/0.78 &0.36/0.75 &\textbf{0.26/0.57} &0.29/0.60 &0.29/0.61\\
					\midrule
					\midrule
					AVG 
					&0.72/1.54 &1.41/2.35 &0.89/1.73 & 0.61/1.21 & 0.54/1.15 &0.48/1.00 &0.48/0.90 &0.43/0.83 &\textbf{0.41/0.82} \\
					\bottomrule
				\end{tabular}
			\end{sc}
        \end{footnotesize}
	\end{center}
	\vskip -0.2in
\end{table*}

\vspace{-1mm}
\section{Experiments}
\vspace{-1mm}
We validate the proposed network on two settings: pedestrian motion prediction, and joint pedestrian and vehicle motion prediction.
\vspace{-1mm}
\subsection{Comparison with SOTAs}
\vspace{-1mm}
\mypar{Metrics} Following previous works~\cite{lee2017desire, alahi2016social, gupta2018social}, we consider two evaluation metrics: average displacement error (ADE)
and final displacement error (FDE). ADE is the average distance between all the predicted trajectory points and the true points, and FDE is the distance between the predicted final destination and the true final destination.

\vspace{-4mm}
\subsubsection{Pedestrian motion prediction}
\vspace{-1mm}
\mypar{Datasets} We consider two public datasets:~\emph{ETH-UCY} and~\emph{Stanford Drone}. ~\emph{ETH-UCY} dataset contains 5 sets, ETH, HOTEL, UNIV, ZARA1 and ZARA2. They consist of human trajectories with rich interactions, such as group forming, dispersing, and collision avoidance. Following the experimental setting in SGAN~\cite{gupta2018social}, we split the trajectories into segments of 8s, where we use 0.4s as the time interval, and use the first 3.2 seconds (8 timestamps) to predict the following 4.8 seconds (12 timestamps). We use the leave-one-out approach, training on 4 sets and testing on the remaining set.~\emph{Stanford Drone} dataset is a crowd pedestrian dataset, including 20 unique scenes on a university campus. The coordinates of multiple actors' trajectories are provided in pixels. Following the standard data-split setting, we use 6356 samples and the same segment split as~\emph{ETH-UCY}.

\mypar{Baselines}
~\emph{SLSTM}~\cite{alahi2016social} pools the hidden states with the neighbors.~\emph{SAT}~\cite{vemula2018social} formulates the trajectory sequence in a spatial-temporal graph to capture the spatial and temporal dynamics.~\emph{CIDNN}~\cite{xu2018encoding} models the crowd interactions with the inner product and introduces a displacement prediction module.~\emph{SGAN}~\cite{gupta2018social} leverages adversarial learning to fit the uncertain human behavior and pools the hidden state with all the other actors involved in the scene. \emph{Sophie}~\cite{sadeghian2019sophie} introduces the attention mechanism discriminatively considering the impact of other actors.~\emph{S-BiGAT}~\cite{kosaraju2019social} introduces GAT to represents the social interactions with a global embedding.~\emph{STGAT}~\cite{huangstgat} captures the temporal interactions with an additional LSTM.~\emph{Desire}~\cite{lee2017desire} takes advantage of the variational auto-encoders and inverse optimal control to generate and rank predictions.~\emph{MATF GAN}~\cite{zhao2019multi} uses shared convolution operations to model the spatial interactions.

\mypar{Results}
Table~\ref{tab:eth-ucy} shows the comparison between the proposed PMP-NMMP against several previous state-of-the-art methods on~\emph{ETH-UCY}. We see that (i) While most of the previous SOTAs~\cite{sadeghian2019sophie,kosaraju2019social,zhao2019multi} are superior on average but perform poorly on some sets, our model is best on 2 sets and compatible on other 3 sets; (ii) PMP-NMMP improves the state-of-the-art to $0.41$m and $0.82$m on ADE and FDE on average. Table~\ref{tab:sdd} shows the performance comparison on~\emph{Stanford Drone}. We see that the proposed PMP-NMMP significantly outperforms the other competitive methods. The intuition is that it uses the NMMP module to effectively capture the social interactions in the crowd scenario.

\begin{table}[h]
	\caption{Quantitative results on~\emph{Stanford Drone} dataset. ADE and FDE are reported in pixels.}
	\label{tab:sdd}
	\vskip 0.1in
	\begin{center}
		\begin{small}
			\begin{sc}
				\begin{tabular}{c|c|c}
					\toprule
					Method & ADE & FDE \\
					\midrule
					SForces &36.38 & 58.14 \\
					SLSTM &31.19 & 56.97 \\
					SGAN &27.25 & 41.44 \\
					Desire &19.25 & 34.05\\
					Sophie &16.27 & 29.38 \\
					MATF GAN &22.59 &33.53 \\
					PMP-NMP &\textbf{14.67} &\textbf{26.72} \\
					\bottomrule
				\end{tabular}
			\end{sc}
		\end{small}
	\end{center}
	\vspace{-5mm}
\end{table}

\begin{figure}[t!]
\begin{center}
\centerline{\includegraphics[width=1.0\linewidth]{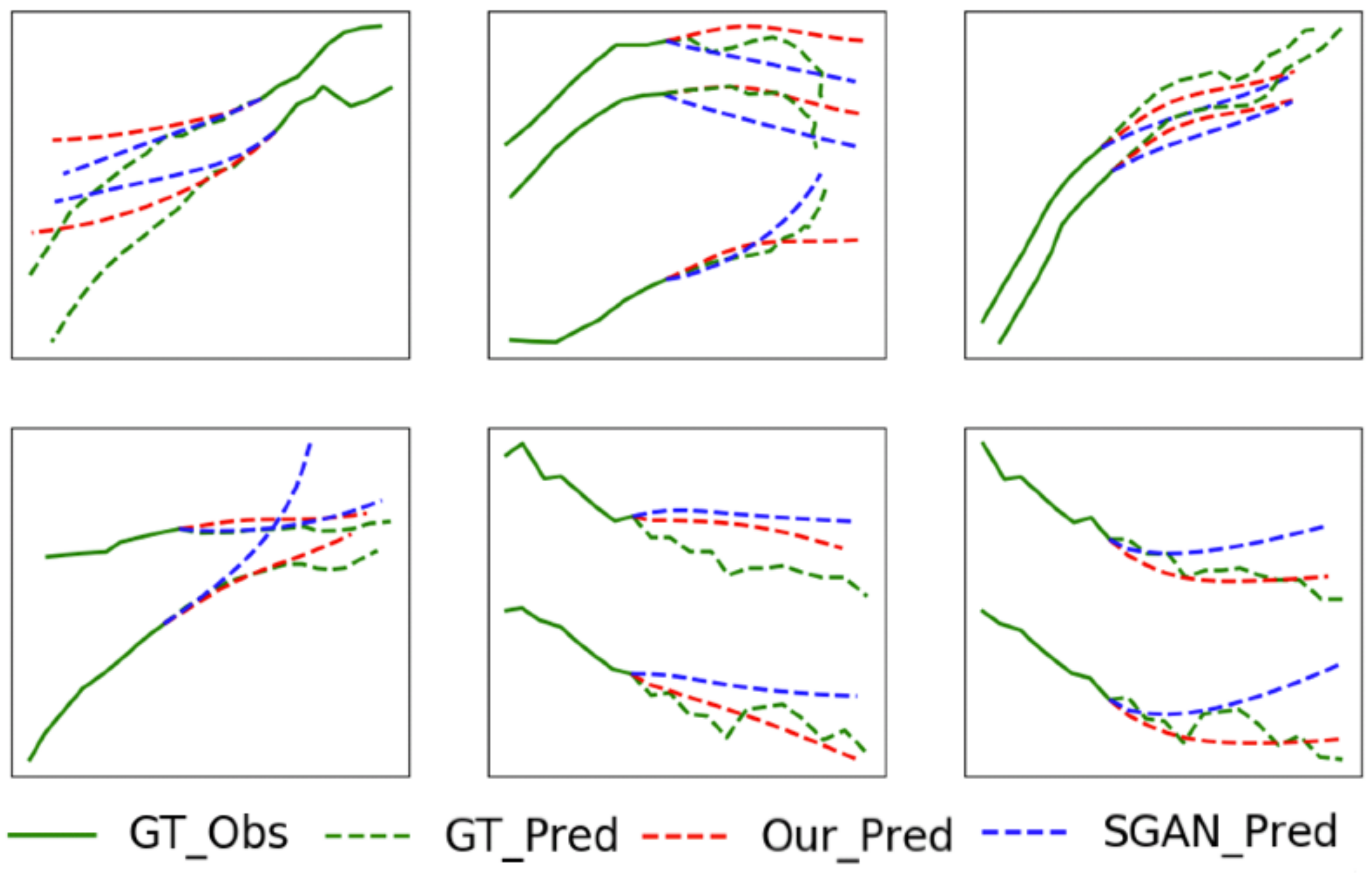}}
\caption{Predicted trajectory comparison in the pedestrian setting. The baseline is SGAN~\cite{gupta2018social}.}
\label{fig:Trajectories}
\end{center}
\vspace{-12mm}
\end{figure}

\mypar{Qualitative comparison}
Figure.~\ref{fig:Trajectories} compares the predicted trajectories to the ground-truth trajectories. We choose six scenarios from HOTEL set and show the ground-truth trajectories (green line), our predictions (dashed red line), and the SGAN baseline (dashed blue line). Our model outperforms SGAN as the corresponding predictions are closer to the ground-truth. 

\vspace{-4mm}
\subsubsection{Joint motion prediction}
\vspace{-2mm}
\mypar{Dataset} We create a joint pedestrian and vehicle motion prediction dataset based on~\emph{NuScenes}~\cite{caesar2019nuscenes}.~\emph{NuScenes} is an autonomous driving dataset, which comprises $1000$ scenes, each $20$s long and fully annotated with 3D bounding boxes. We reorganize the dataset and downsample to avoid the overlapping issue. In total, we select $3148$ samples, where $1888$ for training, $629$ for validation and $631$ for testing.  Each sample contains the ground-truth actors' trajectory information and a 2D scene map. The scene map reflects $100 \times 100$m$^2$ so that 50m in front of and from the back the self-driving vehicle is rasterized. The pixel resolution is $0.2$m and the image size is $500 \times 500$. The time interval is $0.1$s and we forecast the future $3$s (30 timestamps) trajectories from the previous $0.5$s (5 timestamps) trajectories.


\begin{table}[h]
	\caption{Quantitative comparison on~\emph{NuScenes} dataset. Error metrics reported are ADE/FDE in meters. Given previous 5 (0.5s), predicting the future 30 (3s).}
	\label{tab:nuscenes}
	\vspace{-5mm}
	\begin{center}
		\begin{small}
			\begin{sc}
				\begin{tabular}{c|c|c|c}
					\toprule
					Method & Pedestrians & Vehicles & T($ms$)\\
					\midrule
					NoImage & 0.41/0.81 &1.77/3.87 & 1.54 \\
					AlexNet & 0.39/0.79 &1.71/3.79 & 23.3 \\
				 	AlexNet-NMMP &0.38/0.78 &1.59/3.72 &1.98\\
				 	JMP-NMMP-GAN & 0.38/0.78 &1.65/3.81 & 6.23 \\
					JMP-NMMP & {\bf 0.34/0.71} & {\bf 1.54/3.55} &3.76 \\
					\bottomrule
				\end{tabular}
			\end{sc}
		\end{small}
	\end{center}
	\vspace{-5mm}
\end{table}

\mypar{Baselines} ALEXNET~\cite{djuric2018short} extracts the visual cues of the rasterized actor-centric scene image with AlexNet~\cite{krizhevsky2012imagenet} and simultaneously and separately encodes and decodes the prior trajectories with MLPs for each actor in the scene. We exclude the images in ALEXNET, named NOIMAGE. See Figure.~\ref{fig: JointFramework}, we add the NMMP module to ALEXNET and use the SDV-centric scene image instead of multiple actor-centric scene images, named ALEXTNET-NMMP; we substitute the AlexNet with MobileNet-V2~\cite{sandler2018inverted} in ALEXTNET-NMMP, named JMP-NMMP; we further introduce GAN as PMP-NMMP, named JMP-NMMP-GAN.

\mypar{Results} Table~\ref{tab:nuscenes} shows
quantitative comparison. We see that (i) JMP-NMMP consistently performs the best for both vehicles and pedestrians, indicating the superiority of the proposed system; (ii) ALEXNET-NMMP is better than ALEXNET, indicating that the social interactions are crucial when forecasting the future movements; (iii) ALEXNET is better than NOIMAGE and JMP-NMMP is better than ALEXNET-NMMP, indicating the necessity and capacity of scene image; (iv) GAN does not bring gains but add time cost, indicating the traffic actor behaviours are much certain and the GAN interferes the optimization; (v) SDV-centric image highly reduces computational cost and the final running time is sufficiently fast for real-time use.


\begin{figure}[t!]
\vskip 0.1in
\begin{center}
\centerline{\includegraphics[width=0.8\linewidth]{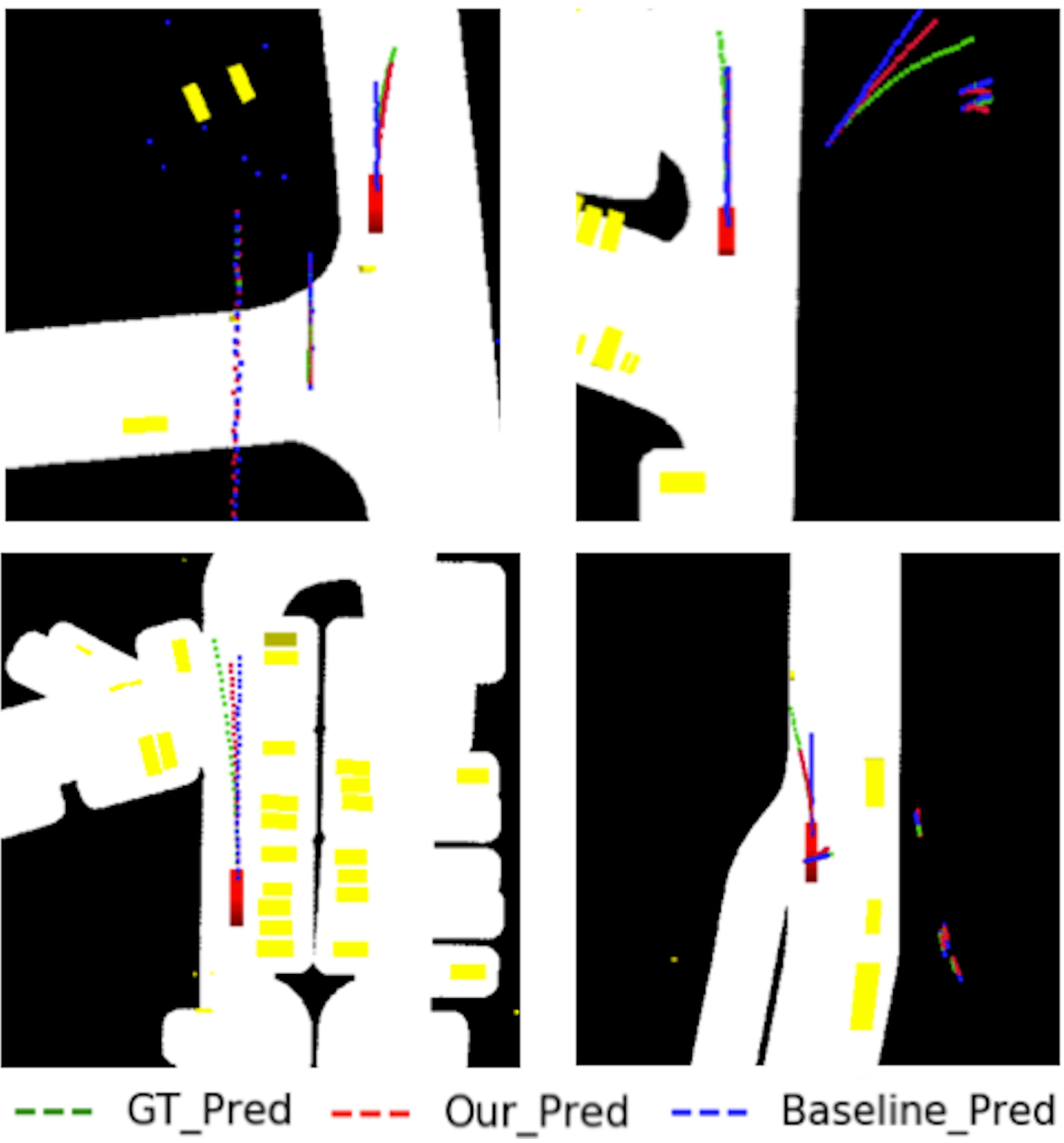}}
\caption{Predicted trajectory comparison in the joint setting. The baseline is ALEXNET~\cite{djuric2018short}.}
\vspace{-12mm}
\label{fig:vehicle_Trajectories}
\end{center}
\end{figure}

\mypar{Qualitative comparison}
Predictions are presented in Figure.~\ref{fig:vehicle_Trajectories}. We choose four samples in the complicated urban street scenarios and only present the moving actors trajectories. Our predictions (red) are more closer to the ground-truth (green) than the baseline (blue) especially in the motion changing cases.

\vspace{-1mm}
\subsection{Ablation study}
\vspace{-1mm}
Since pedestrian motion prediction system and joint motion prediction system are the systems with small variations and have consistent performance, so here we present the analysis of PMP-NMMP as a representative.

\begin{table}[h]
	\caption{We evaluate the NMMP module by applying it in multiple trajectory encoding phases in generator (G) and discriminator (D). \checkmark and - means with and without NMMP module respectively. ADE/FDE are reported in meters.}
	\label{tab:nmp}
	\vspace{-4mm}
	\begin{center}
		\begin{footnotesize}
			\begin{sc}
				\begin{tabular}{c|c|c|c|c|c}
					\toprule
					G D &eth &hotel &univ  &zara1 &zara2\\
					\midrule
					- - &0.74/1.34	&0.50/1.01	&0.64/1.29	
			   &0.34/0.69	&0.35/0.73 \\
                    \checkmark - &0.70/1.30	&0.41/0.80	&0.54/1.15
               &0.33/0.68	&0.31/0.65\\
                    \checkmark \checkmark &{\bf 0.61/1.08}	&{\bf 0.33/0.63}	&{\bf 0.52/1.1}
               &{\bf 0.32/0.66} & {\bf 0.29/0.61} \\
					\bottomrule
				\end{tabular}
			\end{sc}
		\end{footnotesize}
	\end{center}
	\vspace{-5mm}
\end{table}

\mypar{Effect of NMMP module in PMP-NMMP} 
To better model the social interactions between actors involved in the scene, we propose the NMMP module. In PMP-NMMP, there are two trajectory encoding phases that use the NMMP module: one in the generator and the other one in the discriminator. Table~\ref{tab:nmp} compares three models. The first one is the baseline, where we adopt the pooling module in~\cite{gupta2018social} instead of the NMMP module to model the social interactions. The second one substitutes the pooling module in the generator with the NMMP module. The third one further substitutes the pooling module in the discriminator with our NMMP module. The stable gains on all the 5 sets shows the strength of the NMMP module in modeling the interactions.

\begin{table}[t!]
	\caption{The performance as a function of the number of iterations in NMMP ($K$) on UNIV set.}
	\label{tab:nmp layer}
	\vspace{-2mm}
	\begin{center}
		\begin{small}
			\begin{sc}
				\begin{tabular}{c|cccc}
					\toprule
					$K$  & 1 & 3 & 5 & 7 \\
					\midrule
					ADE & 0.55 & 0.54 & {\bf 0.52} & 0.56 
					\\
					FDE & 1.16 & 1.16 & {\bf 1.11} & 1.20 \\
					\bottomrule
				\end{tabular}
			\end{sc}
		\end{small}
	\end{center}
	\vspace{-8mm}
\end{table}

\mypar{Effect of the number of iterations in NMMP}
We explore the performance of NMMP with different number of iterations on the most crowded set (UNIV), where pedestrian amounts ranges from $2$ to $57$ in a single scene. Table~\ref{tab:nmp layer} shows the quantitative results. The performance gets better, and then sharply worse with the increasing number of iterations. The intuition is as follows. With the increasing number of iterations, the NMMP module gets more capacity to model the dense interactions; however, when there are too many iterations, the actor embedding would be over-smoothing, mixing local information and global context.

\begin{table}[h]
	\caption{The exploration of the decoder design on the UNIV set. ADE/FDE are reported in meters.}
	\label{tab:decoder structure}
	\vspace{-2mm}
	\begin{center}
		\begin{small}
			\begin{sc}
				\begin{tabular}{c|cc|c}
					\toprule
				& Individual & Final	& ADE/FDE \\
					\midrule
					Single &  & \checkmark &0.56/1.19 \\
					Double &  & \checkmark &{ \bf 0.52/1.11} \\
					Double & \checkmark &  &0.76/1.17 \\
					\bottomrule
				\end{tabular}
			\end{sc}
		\end{small}
	\end{center}
	\vspace{-4mm}
\end{table}

\mypar{Effect of decoder structure in PMP-NMMP}
We explore the design of the decoder structure from two aspects. The first aspect compares the single-decoder structure and the double-decoder structure. The single-decoder structure fuses the individual trajectory embedding and interacted actor embedding in the feature domain, and then feeds the unifying features to a single decoder to predict the final trajectory; the double structure apply the individual and interactive decoders to generate two outputs, which add up to the final trajectory.
The comparison between the first and second rows in Table~\ref{tab:decoder structure} shows that the double-decoder structure outperforms the single-decoder structure, indicating the double-decoder structure can better model the impacts from other actors involved in the scene. 
The second aspect compares the output of the individual branch and the final output in the double-decoder structure.
The comparison between the second and third rows in Table~\ref{tab:decoder structure} shows that the final output outperforms the output of the individual branch, indicating the importance of the interactive branch.

\begin{figure}[!b]
\vskip -0.2in
\begin{center}
\centerline{\includegraphics[width=0.95\linewidth]{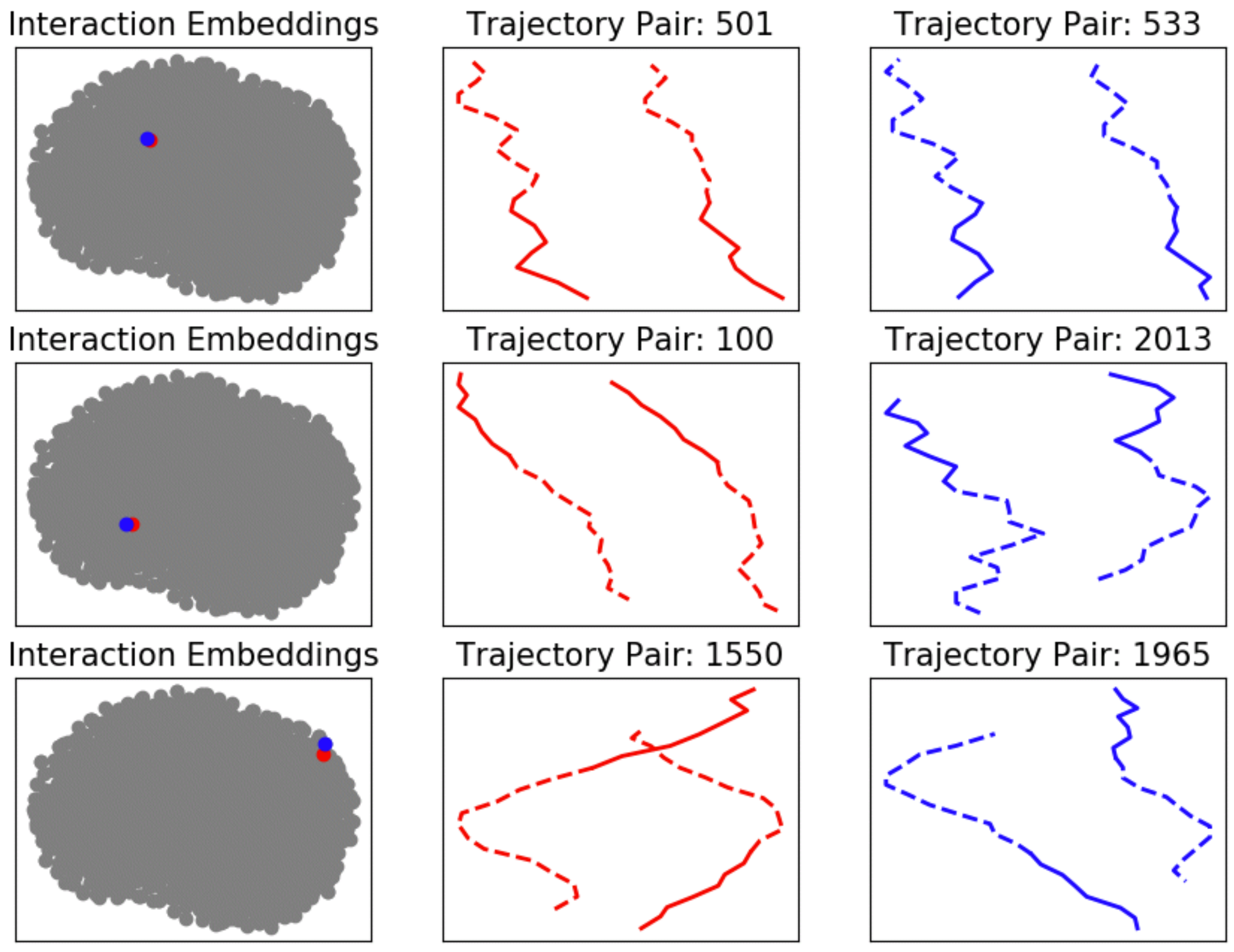}}
\caption{Close interaction embeddings in the embedding domain lead to similar trajectory relations between the corresponding actor pairs in the spatial domain. The first column is the visualization of interaction embeddings, where each dot represents an interaction and corresponds to one spatial-domain plot in the right columns.}
\label{fig:t-SNE}
\end{center}
\vspace{-10mm}
\end{figure}

\vspace{-2mm}
\subsection{Interpretability exploration}

\mypar{Visualization of interaction embeddings in NMMP}  Figure~\ref{fig:t-SNE} shows the visualization of interactions in the embedding domain and the corresponding actor pairs in the spatial domain  in HOTEL set. The interaction embeddings $\mathbf{e}_{ij}$ are mapped to 2D coordinates via t-SNE and shown in the first column. We randomly pick three pairs of close interaction samples in the embedding space, which are colored blue and red, and plot the corresponding trajectory pairs in the followed two columns. The full line denotes the observed trajectories, while the dash line represents the future trajectories. 
We see that (i) close interaction embeddings in the embedding domain lead to similar trajectory relations between associated actors in the spatial domain. For example, in the first row, pedestrians are walking upwards in parallel; in the third row, pedestrians are walking toward opposite directions; (ii) different interaction types between actors' trajectory in the spatial domain are encoded to different positions in the embedding domain. The top two rows are close; both are far from the last row in embedding space. The trajectories show the interaction types of the top two rows are similar and quite different from the last row.

\mypar{Performances in crowd scenarios} 
We split 5 sets of ETH-UCY dataset according to the pedestrian amounts in the scene. We evaluate the proposed PMP-NMMP and the SGAN~\cite{gupta2018social} baseline on this re-split dataset. Then we count the corresponding ADE/FDE normalization ratio
\vspace{-1mm}
$$
r_{\rm ADE/FDE} = \frac{{\rm SGAN}_{\rm ADE/FDE} -{\rm NMMP}_{\rm ADE/FDE} }{{\rm SGAN}_{\rm ADE/FDE} }
$$ 
for the pedestrians in each split. A positive $r_{\rm ADE/FDE}$ indicates that PMP-NMMP outperforms SGAN.

\begin{figure}[t!]
\begin{center}
\centerline{\includegraphics[width=0.95\linewidth]{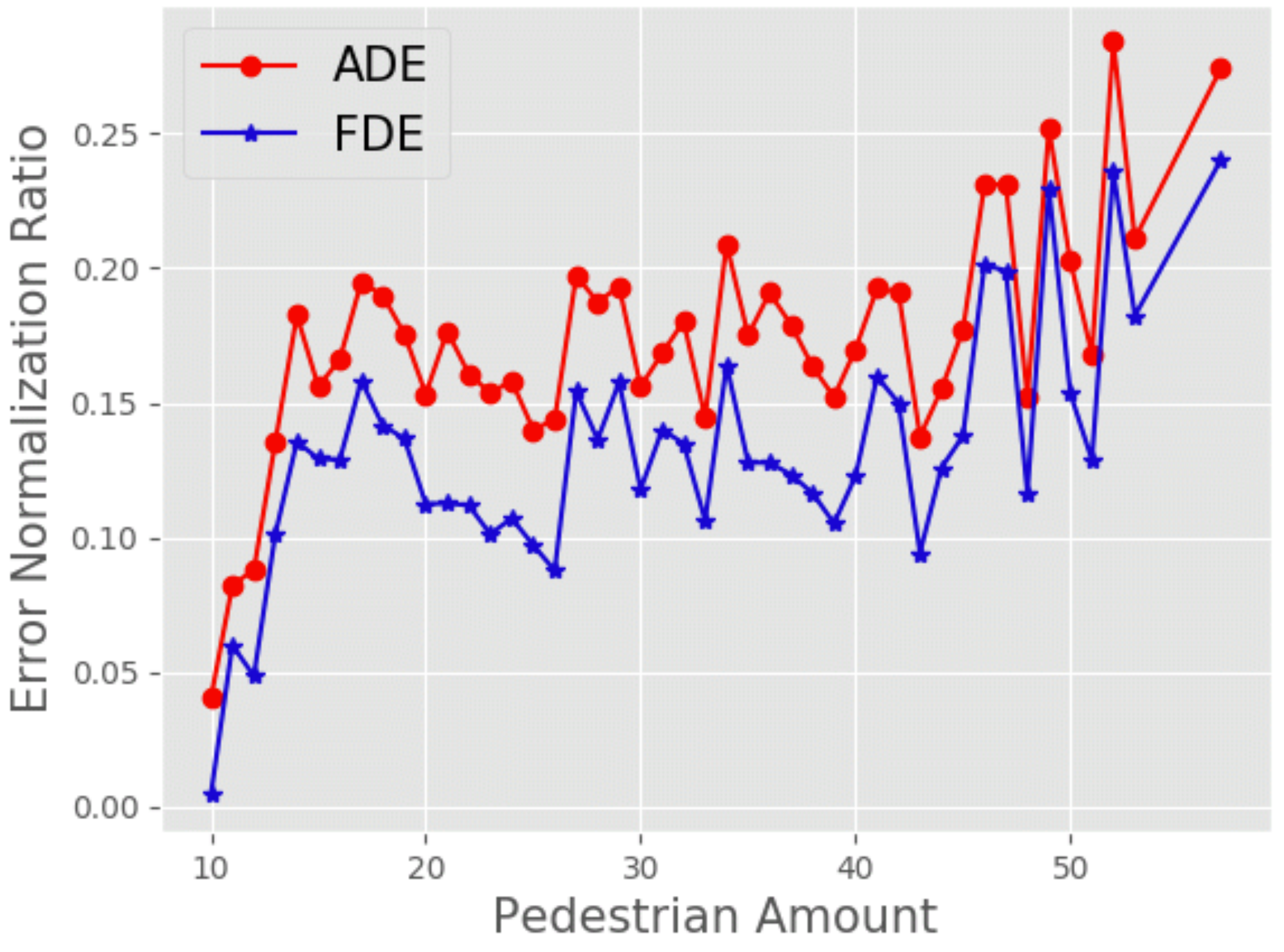}}
\caption{The ADE and FDE normalization ratio over pedestrian amounts in the scene. The $x$-axis is the total pedestrian amount in the scene. The $y$-axis is the ADE (Red) and FDE (Blue) normalization ratio compared with SGAN~\cite{gupta2018social}.}
\label{fig:ErrorReduce}
\end{center}
\vspace{-10mm}
\end{figure}

Figure~\ref{fig:ErrorReduce} shows the ADE/FDE normalization ratio as a function of the number of pedestrian in the scene. We see that (i) the normalization ratio is always positive, indicating PMP-NMMP consistently outperforms SGAN;  and (ii) the normalization ratio increases as  the numbers of pedestrian gets larger, indicating the advantage of NMMP over 
SGAN is larger. The intution behind is that the NMMP module has a larger capacity to handle the crowd scenarios, where the social interactions are abundant.

\vspace{-2mm}
\section{Conclusions}
\vspace{-2mm}
In this work, we propose NMMP, a novel module explicitly modeling the interactions between the traffic actors in a scene. Unlike prior work simply tackling a single motion prediction task, we design NMMP based systems for pedestrian motion prediction, and joint pedestrian and vehicle motion prediction. Both outperforms prior state-of-the-art methods across several widely used benchmarks.
Through our evaluations and visualizations, we show that NMMP is able to capture the interaction patterns and our systems can predict precise trajectories in different scenarios.

\vspace{2mm}
\noindent
\textbf{Acknowledgements} 
This work is supported by the National Key Research and Development Program of China (2019YFB1804304), NSFC (61521062), 111 plan (B07022), and STCSM (18DZ2270700).

{\small
\bibliographystyle{ieee_fullname}
\bibliography{ms}
}

\newpage
\begin{Large}
\noindent
\textbf{Further experimental analysis}
\end{Large}
\vspace{2mm}

\mypar{Rasterization}
Given the global coordinates and the headings of each traffic actor involved in the scene and the map marked the drivable areas. We encode each actor type into the corresponding vector layer. White represents drivable areas, and vice versa. Vehicles are in yellow polygons while the self-driving vehicle (SDV) is in red and pedestrians in green. Other actors are colored black. Besides, the actors' histories are represented with reduced level of brightness in the same color. The vector layers of the same actor are rasterized one by one in time order, resulting in the fading effect. Then the vector layers are rasterized one by one on top of each other, in the order from drivable areas to traffic actors such as vehicles and pedestrians, resulting in a RGB scene image.

\begin{figure}[!b]
\vskip -0.1in
\begin{center}
\centerline{\includegraphics[width=0.95\linewidth]{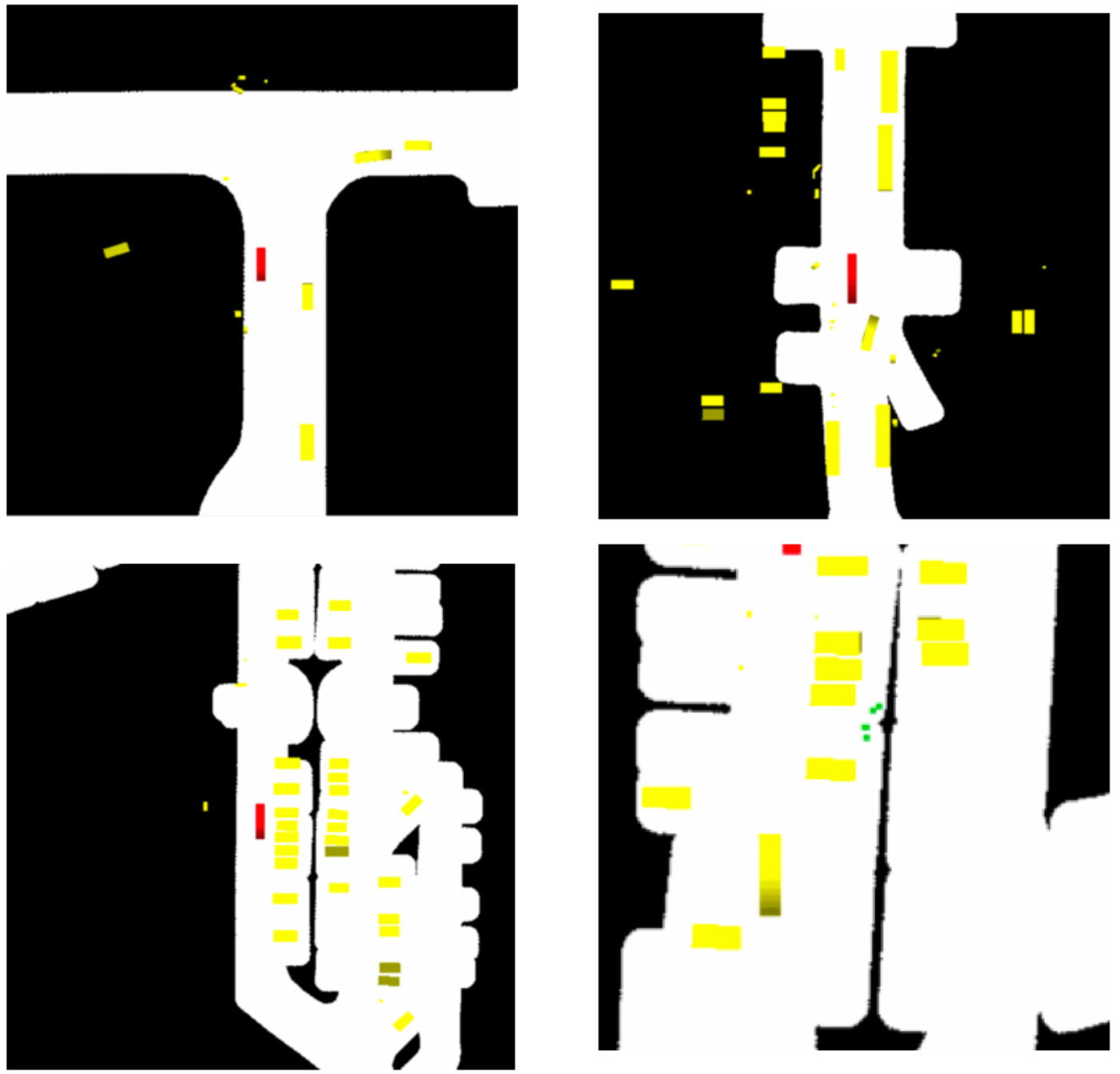}}
\caption{Rasterized scene images. The red box is the SDV, the yellow boxes are the vehicles, and the green dots are the pedestrians. As time goes back, the brightness of the boxes darkens, resulting in the fading tails.}
\label{fig:t-SNE}
\end{center}
\end{figure}

\begin{figure}[!t]
\begin{center}
\centerline{\includegraphics[width=1.0\linewidth]{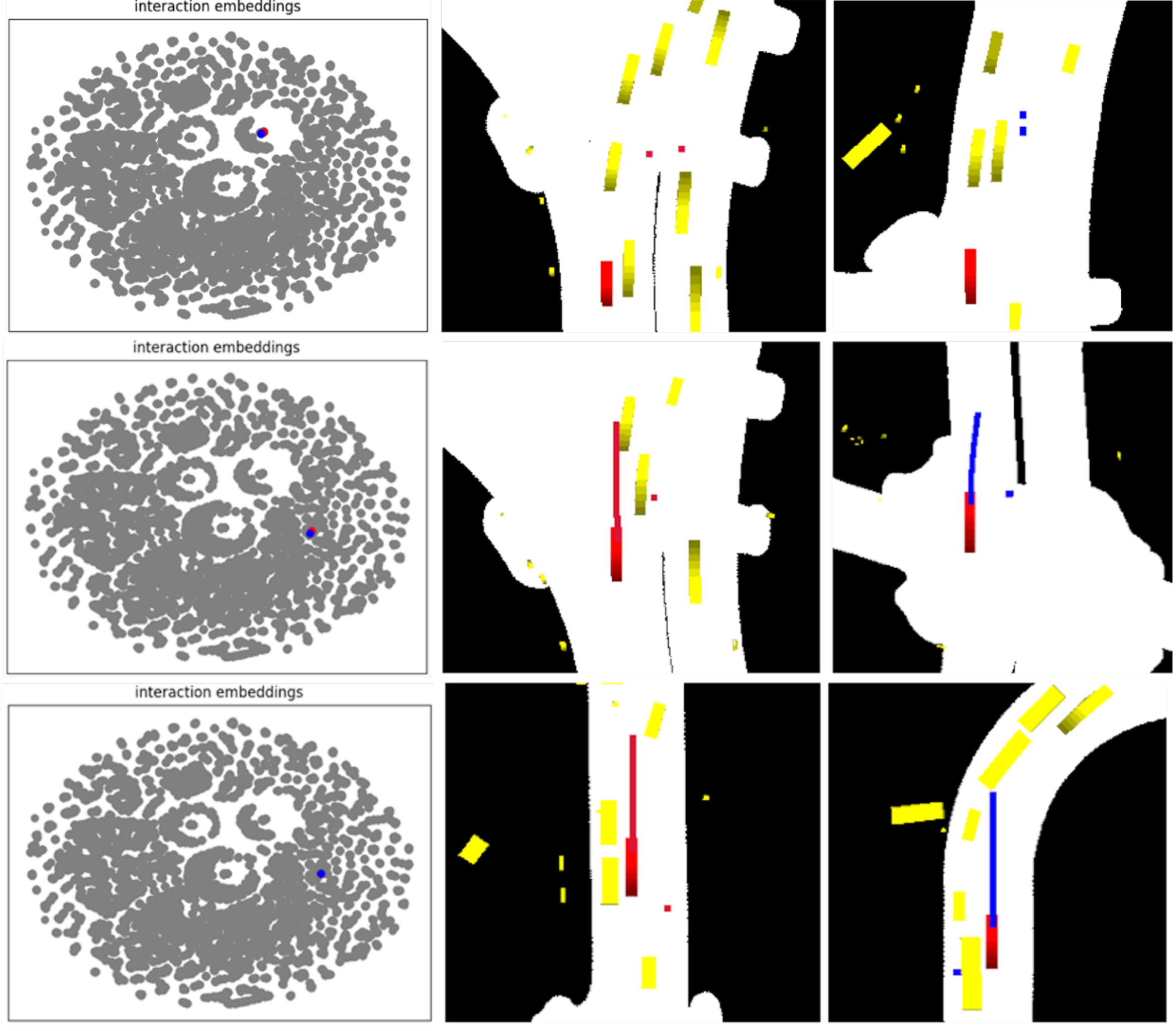}}
\caption{The first column is the t-SNE visualization of interaction embeddings, where each dot represents an interaction and correspond to spatial-domain plots. The close red and blue dots in the first column represent close interaction embeddings in the embedding domain. The right two columns are the corresponding traffic scenes. And the middle column is the corresponding spatial plot of the red dot. The trajectories of the interacted traffic actors are colored in red. Similarly the right column presents spatial plot of the blue dot.}
\label{fig:t-SNE-JMP}
\end{center}
\vskip -0.3in
\end{figure}


\mypar{Visualization of interaction embeddings in JMP-NMMP} Figure.~\ref{fig:t-SNE-JMP} shows the visualization of interactions in the embedding domain and the corresponding actor pairs in the spatial domain in the joint motion prediction scenarios. Similar to interaction embedding visualization in Figure~\ref{fig:t-SNE}, the interaction embeddings $\mathbf{e}_{ij}$ are mapped to 2D coordinates via t-SNE and shown in the first column. We randomly pick three pairs of close interaction samples in the embedding space, which are colored red and blue, and plot the corresponding trajectory pairs in the followed two columns with dots. We get the similar conclusions as the interaction embedding analysis in pedestrian motion prediction setting. We see that (i) close interaction embeddings in the embedding domain lead to similar trajectory relations between associated actors in the spatial domain. For example, in the first row, pedestrians are following the front one and moving slowly to pass the road; in the second and third rows, pedestrians are obstructed by the moving vehicles on the road; (ii) different interaction types between actors' trajectory in the spatial domain are encoded to different positions in the embedding domain. The bottom two rows are close; both are far from the top row in embedding space. The trajectories show the interaction types of the top two rows are similar and quite different from the last row.

\end{document}